\documentclass[runningheads,a4paper]{llncs}
\usepackage{tikz}
\usepackage{todonotes}
\usepackage{mathtools}
\usepackage{wrapfig}
\usepackage{graphicx}
\usepackage{epstopdf}
\usepackage{enumitem}
\usepackage{url}
\usepackage{xcolor}
\usepackage{verbatim}
\usepackage{verbatimbox}
\usepackage{algorithm}
\usepackage[noend]{algpseudocode}
\usepackage{mathrsfs}  
\usepackage{amssymb,amsmath,amsfonts} 
\usepackage{caption}
\usepackage{subcaption}
\usepackage{pgfplots}
\usepackage{soul}

  \pgfplotsset{
       compat=1.11,
       every axis/.append style={       
       legend style={font=\fontsize{16}{0}\selectfont},
       tick label style={font=\fontsize{16}{0}\selectfont},       
       x label style={scale=2.5},
       y label style={scale=2.5},
},
    }    
\usepackage{pgfplotstable}
\urldef{\mailsa}\path|name.Name@place.ac.uk| 

\newcommand \Prob {\mathbb{P}}

\newcommand {\pd}[1]{p\left(#1\right)}
\newcommand{\Tp}{\mathbb{T}}
\newcommand {\T} {\Tp}
\newcommand \param {\theta}

\newcommand \Param {\Theta}
\newcommand \Dt {D}
\newcommand \D {D}
\newcommand \prop {\phi}
\newcommand \pathprop {\varphi}
\newcommand \Act {\operatorname{Act}}
\newcommand \Props {\mathbf{P}}
\newcommand{\M}{\mathbf{M}}
\newcommand{\Mp}{{\mathbf{M}(\param)}}
\newcommand{\Mset}{\mathbf{M}_\Theta}
\newcommand{\s}{\mathbf{S}}
\newcommand {\sys} {\s}

\newcommand{\strategy}{\pi}

\newcommand{\EX}[1]{\mathbb{E}\left(#1\right)}
\newcommand{\EXsa}[1]{\mathbb{E}_{s,\alpha}\left(#1\right)}
\newcommand{\Pred}[1]{#1^{\operatorname{pred}}}%{E\left(#1\right)}
\newcommand \conf {\mathcal{C}}
\newcommand \Gain {\mathbb{G}}

\DeclareMathOperator{\Dir}{\operatorname{Dir}}

\newcommand{\AP}{\operatorname{AP}}

\tikzset{
    blob/.style={
           circle, 
           draw=black,
           inner sep=1pt,
           minimum size=1.5em,
           text width=1.2em,
           text centered,
           },
               newblob/.style={
           circle, font=\small,
           draw=black, fill=black!5,
        inner sep=0.8pt,
           minimum size=0.5em,
           text width=0.8em,
           text centered,
           },
}

	\newtheorem{defn}[definition]{Definition}

\DeclareCaptionFont{mysize1}{\fontsize{5}{0}\selectfont}
\DeclareCaptionFont{mysize2}{\fontsize{8}{0}\selectfont}
\usetikzlibrary{automata,arrows, positioning, intersections, shapes, calc, angles, quotes}

\tikzstyle{vertex}=[draw,fill=black!15,circle,minimum size=20pt,inner sep=0pt]
\begin{document}
\mainmatter  
\title{Automated Experiment Design\\ for Data-Efficient Verification\\ of Parametric Markov Decision Processes}
\titlerunning{Automated Experiment Design for Data-Efficient Verification of pMDPs}
\author{E.~Polgreen\inst{1} \and V.B.~Wijesuriya\inst{1}\and S.~Haesaert\inst{2}  \and A.~Abate\inst{1}
}
\authorrunning{E.~Polgreen et al.}

\institute{
Department of Computer Science,  
University of Oxford \and% 
Department of Electrical Engineering,  
Eindhoven University of Technology} 
\maketitle

\begin{abstract}
% To much commented uncommented, I just re-started based on the original text.
We present a new method for statistical verification of quantitative properties over a partially unknown system with actions, 
utilising a parameterised model (in this work, a parametric Markov decision process)   
and data collected from experiments performed on the underlying system.   
% For systems that can be modelled by parametric Markov decision processes, 
We obtain the confidence that the underlying system satisfies a given property,  
and show that the method uses data efficiently and thus is robust to the amount of data available.  
These characteristics are achieved 
%\ale{[we do not really mention parameter synthesis in the next two points, which both refer to data]} 
by firstly exploiting parameter synthesis to establish a feasible set of parameters for which the underlying
system will satisfy the property;
% shaping the parameterised model around the property of interest; 
secondly, by actively synthesising experiments to increase amount of information in the collected data that is relevant to the property;
and finally propagating this information over the model parameters, obtaining a confidence that reflects our belief whether or not the system 
parameters lie in the feasible set, thereby solving the verification problem.  
%Beyond extending statistical verification to partial probabilistic models with actions and nondeterminism, 
%unlike current techniques that are limited to bounded-time properties,  
%by combining parameter synthesis with statistical techniques 
%our method can tackle both bounded and unbounded-time properties in a data-efficient manner.  
%Experimental evidence 
%We implement our method and evaluate its data-efficiency when verifying 
% an unbounded-time fragment of PCTL over partial probabilistic models with actions and nondeterminism. 
\end{abstract}

%%%%%%%%%%%%%%%%%%%%%%%%%%%%%%%%%%%%%%%%%%%%
%Section 1: Introduction
%%%%%%%%%%%%%%%%%%%%%%%%%%%%%%%%%%%%%%%%%%%%%
%!TEX root = MDPpaper.tex
%
\section{Introduction}

%\ale{[fix capitalisation throughout article for: \\ 
%Markov decision process, MDP \\ 
%parametric Markov decision process, pMDP ]}

Formal verification 
%is a powerful tool for guaranteeing correctness of complex systems. These guarantees correctness, however, 
relies on full access to accurate models describing the behaviour of systems in order to guarantee their correctness.  
Such models are often hard to obtain for systems encompassing partially understood behaviours and uncertain events. 
For a partially unknown system, the unknown model characteristics
can be represented via non-determinism in the form of parameters. 
The resulting parameterised model captures all available knowledge on the underlying system of interest. 

We target the verification of a fragment of Probabilistic Computation Tree Logic (PCTL) on partially unknown systems with actions. 
We develop a new approach that incorporates the available information captured by a parameterised model with the active collection of a limited amount of data from the underlying system. 
The verification problem is tackled in three phases. %Extending an earlier contribution \cite{DBLP:conf/qest/PolgreenWHA16}, 
In the first phase we use the available parameterised model to synthesise the %feasible
 set of parameters for which the property of interest is satisfied (called the \textit{feasible set}).
 
In the second phase 
a series of experiments are designed and executed on the system to 
update the knowledge available about the parameters of the parameterised model.  
More precisely, 
a procedure executes the designed experiments, 
obtains data from the system and, 
by means of Bayesian statistics, 
updates distributions over the likely parameter values of the parameterised model. 
This updated knowledge is returned to the experiment design module, 
and the process is repeated until a preset limit on the total amount of collectible data is reached. 
The design of %ing of such 
such experiments is important to attain a reasonable level of confidence in the acceptance or rejection of a property with %increased accuracy and 
a limited amount of data. 

%The second phase forms a loop. 
In the final phase, we combine the output from the parameter synthesis
with the updated distributions over the model parameters
to quantify the confidence that the system satisfies (or does not satisfy) the property.  

This work extends the contributions in~\cite{DBLP:conf/qest/PolgreenWHA16} by focussing on systems with actions: 
the presence of (action) non-determinism provides the potential for experiment design, 
whereby we select actions to improve the accuracy of our confidence value. 
More precisely 
we design experiments that maximise the usefulness of the data collected. 
Intuitively, 
this means that we want to design experiments to prioritise the collection of data that leads to proving or disproving the satisfaction of the property. 
%affects
%As detailed later, this experiment design problem differs from related literature on \emph{active learning}. 
%A key new contribution of our experiment design is that, by combination with parameter synthesis
%on the partial model, we can give greater importance to learning parameters that 
%contribute significantly towards satisfaction of the property.
%\ale{[The remaining part of this important paragraph is not particularly clear, as it does not 
%elaborate on why our problem is different and more difficult/interesting. ]}
%In the active learning of labelled Markov decision processes~\cite{DBLP:conf/icmla/ChenN12}, 
%the utility is simply expressed via a Kullback-Leibler divergence; 
%in reward-based problems such as Bayesian Reinforcement Learning, 
%synthesis is tackled with belief-dependent rewards in~\cite{DBLP:conf/ewrl/Araya-LopezBTC11}. 
In this work, we present the complete approach, 
and evaluate the contribution of our experiment design procedure .  
%towards improvement in accuracy of the confidence assertion. 
We argue that \textit{automated experiment design} allows us to draw sensible conclusions robustly with a limited amount of data.  

\vspace{-0.5cm}
\subsubsection{Structure of the paper.}
Section~\ref{sec:background} provides the necessary background information for the rest of paper to build upon. 
Section~\ref{sec:overview} presents an overview of our algorithm. 
Subsequent sections detail the different phases of the algorithm: 
in Section~\ref{sec:inf} we show how we collect data; 
Section~\ref{sec:conf} provides details on the confidence computation;
and the key contribution of this work is Section~\ref{sec:strat_synth}, 
which outlines our experiment design approach. 

\subsection{Problem Statement}

Consider a partially unknown %dynamical 
system $\mathbf S$, with
external non-determinism in the form of actions, and suppose we can
gather a limited amount of sample trajectories from this system.
 Assume the partial knowledge about the system is encompassed within a
parameterised model class describing the behaviour of $\mathbf S$. %, underall possible sets of actions,
 %up to the unknown parameterisation of some
%of its transitions. 
We investigate two sub-problems: \\[-1.5em]
\begin{itemize} 
\item Can we efficiently use this limited amount of data from a system $\mathbf S$ 
%Can we efficiently use the gathered data 
%and the model knowledge of $\mathbf S$ 
to 
quantify a confidence that the system $\sys$ verifies a given PCTL property? 
\item How should we design an experiment on the system such that the gathered data allows us to verify the property with the greatest degree of accuracy? 
Let the choice of actions of system $\mathbf S$ be something we can control during the experiment,  
and let there only be a limited amount of available experiment time;
can we optimise the sequence of actions to increase the accuracy of the confidence quantification?  
%for a given amount of data? 

\end{itemize}

\subsection{Related Work}
 
%{\color{red}[Use your literature review to defend and prove each of these statements. ]} 
 
%Key benefits of the new approach are: 
%\begin{itemize}
%\item 
% Our method can tackle properties that \ale{[undefined acronym, not discussed earlier]} SMC for MDP cannot, namely unbounded time properties.
%\item  
% Our method combines partial model knowledge with data-based learning.  
% This allows us to give a confidence in whether the system satisfies a given property (or not), 
% which is not available from approaches that separately learn the model and then use conventional verification techniques. 
%\item  
% Our method exploits partial model knowledge, 
% in order to use data more efficiently than existing methods for learning MDP, 
% \ale{because we allow linear relations between transition probabilities.} 
%\item  
%% Our proposed algorithms for choosing input actions to the system, 
 %unlikely the active learning methods in~\cite{DBLP:conf/icmla/ChenN12,DBLP:conf/ewrl/Araya-LopezBTC11}, 
% account for the importance of unknown transition probabilities for the satisfaction of the specified property.
% We trade off computational efficiency with data efficiency, 
% and we show that a data-efficient 
 %but computationally-intensive 
% method produces more accurate confidence values than randomised input actions, 
% at the cost of synthesising optimal actions.  
%\end{itemize}  

We compare our work to two branches of research: Statistical Model Checking (SMC) and
research concerned with learning models from system data. We contrast our experiment design method
with existing strategy synthesis techniques for \textit{fully known} Markov decision processes (MDPs).

We emphasise that we tackle a different problem than SMC: 
we target partially unknown systems and gather data from the underlying system; 
SMC\cite{DBLP:conf/rv/LegayDB10} targets fully known models that 
are too big for conventional verification, 
and generates large amounts of data from the models themselves. 
When applied to model-free scenarios~\cite{Sen2004,DBLP:conf/cav/Younes05},
 SMC generates this data from the underlying system. 
By using partial model knowledge, we substantially reduce this data requirement. 
In addition, SMC for systems with non-determinism~\cite{DBLP:journals/sttt/DArgenioLST15,DBLP:conf/qest/HenriquesMZPC12} 
considers only bounded-time properties, and depends on the ability
to generate traces from the model of length greater than the bound. 
By incorporating parameter synthesis tools, 
we are able to consider unbounded-time properties and to draw conclusions from much shorter traces.  

Research on learning models from system data is broad. 
\cite{Eichelsbacher2002,DBLP:journals/jmlr/RossPCK11} use a Bayesian approach to learn full
 Markov models of completely unknown systems. Our work uses a similar Bayesian method but 
 differs because we include information from
 the partial model, which allows us to consider known relationships between parameters
 and thus reduce the amount of data needed for inference.
%\ale{however our consideration of a partial model would allow one to include relationships between parameters and in comparison to substantially reduce the data needed. }
\cite{DBLP:conf/ewrl/Araya-LopezBTC11,DBLP:conf/icmla/ChenN12} use active learning 
%(or experiment design) 
to discover full MDP models from data, prioritising actions by\textit{ variance minimisation} or \textit{KL divergence}. 
The inclusion of a partial model in our method allows us to instead
prioritise gathering data that contributes to the acceptance or rejection of a given property over the system. 
Although~\cite{DBLP:conf/icmla/ChenN12} learns the model with the goal of system verification, 
the authors provide no means of quantifying a confidence that the system satisfies the property, as they do not
have a way to assess which transition probabilities have the greatest contribution to the satisfaction of the property.  

Considering different model classes, experiment design is also used in system identification~\cite{A-GevBomHilSol11}. 
Recent studies~\cite{haesaert2015data,7810322} have incorporated experiment design to data-driven statistical 
verification over \textit{dynamical systems} with partly unknown dynamics, controllable inputs and noisy measurements. 
Similar to our approach, they also compute a confidence estimate on the properties of interest by gathering data through \textit{optimal experiment design}. 

Action selection for Markov decision processes, though
in our context used for experiment design,  
is a known problem that in general amounts to synthesising strategies. 
\cite{DBLP:conf/atva/KwiatkowskaP13} presents an overview for MDPs with static rewards, 
and~\cite{DBLP:conf/uai/GrettonPT03} provides solutions for MDPs with non-Markovian rewards. 
Closer to our approach,~\cite{guan2014online} synthesises strategies for MDPs online,  
where an agent learns a state cost only after selecting an action. 
\cite{5658} use inference-based techniques over strategies to pick a strategy that maximises the expected reward for an MDP with arbitrary rewards.

%%%%%%%%%%%%%%%%%%%%%%%%%%%%%%%%%%%%%%%%%%%%
%%%%%%%%%%%%%%%%%%%%%%%%%%%%%%%%%%%%%%%%%%%%
%Section 2: Background
%%%%%%%%%%%%%%%%%%%%%%%%%%%%%%%%%%%%%%%%%%%%
% !TEX root = MDPpaper.tex
\section{Background}%:Background}
\label{sec:background}

We model a fully known system as a Markov decision process~\cite{Baier2008}. 
%\brian{A Markov decision process 
%can be considered as an adaptation 
%%of a Markov chain permitting nondeterministic
%specifications incorporating actions, as given below}.
%defined as follows.

\begin{defn} \label{def:MDP}
A discrete-time Markov decision process (MDP) $\M$ is a tuple $(S, \Act, \Tp, \iota_{init}, \AP,
L)$, where:
\begin{itemize} 
\item $S$ is a finite, non-empty set of states, 
\item $\Act$ is a set of actions, 
\item $\Tp : S \times Act \times S \to [0,1]$ is the transition probability function, 
such that $\forall s\in S$ and $\forall \alpha \in \Act$, $\sum_{s'\in S}\Tp(s, \alpha, s') \in \{0,1 \}$, 
\item  $\iota_{init}: S \to[0,1]$ denotes an initial probability distribution over the states $S$,
such that $\sum_{s \in S} \iota_{init}(s) = 1$,   
\item The states in $S$  are labelled with  atomic propositions  $a\in \AP$ via the labelling
function $L: S \to 2^{{\AP}}$.
\end{itemize}
An action $\alpha\in Act$ is enabled in state $s$ if and only if $\sum_{s'\in S}\Tp(s, \alpha, s') =
1$. Let $\Act(s)$ denote the set of enabled actions in $s$. For any state $s \in S$, it is required
that $\Act(s) \neq \varnothing$. Each state $s'\in S$ for which $\Tp(s, \alpha, s')>0$ is called an
$\alpha$-successor of $s$. 
Those states $s$ satisfying the condition $\iota_{init}(s)>0$ are called initial states.
\end{defn}
We assume that the MDP is not known exactly, and instead belongs to the set of MDPs represented by a parametric Markov decision process. %
%We assume that the underlying system  belongs to the set of MDPs induced by a given parametric Markov decision process as defined by our model. 

\begin{defn}
\label{def:pMDP}
A discrete-time parametric Markov decision process (pMDP) is a tuple $\Mset= (S, \Act,
\Tp_\param, \iota_{init}, \AP, L, \Param)$, where $S, \iota_{init}, \Act, \AP, L$ are as in
Definition \ref{def:MDP}. 
The entries in $\Tp_\param$ are specified in terms of parameters, collected in a parameter vector
$\param \in \Param$, 
where $\Param$ is the set of all possible evaluations of $\param$. 
Each evaluation gives rise to an induced Markov decision process $\Mp$.
\end{defn}
$\forall s\in S, \forall \alpha \in \Act(s), \forall \param\in\Param: \sum_{s'\in S}\Tp_\param(s,\alpha, s') =
1$, 
namely any $\param \in \Param$ induces an MDP $\Mp$ where the
transition function $\Tp_\param$ can be represented by a \textit{stochastic matrix}. We also assume a
prior distribution on the model parameters (to be used in Bayesian inference). We assume all non-parameterised transition probabilities are known exactly.

As in~\cite{DBLP:conf/qest/PolgreenWHA16}, we consider \textit{linearly parameterised} MPDs,  
where unknown transition
probabilities can be linearly related. 
More precisely, 
given $\Param\subseteq  [0,1]^{n}$ and parameter vector $\param = (\param_1,\ldots,\param_n)\in
\Param$ with $\param_i\in [0,1]$, a pMDP is considered linearly parameterised 
if all outgoing transition probabilities of state-actions pairs have probability $g_l(\param)$ or  $1-g_l(\param)$, 
where  $g_l(\param) = k_0 + k_1\param_1 + ... + k_n \param_n$ with
$k_i\in[0,1]$ and $\sum k_i \leq 1$. This restriction is due to the transformations presented in~\cite{DBLP:conf/qest/PolgreenWHA16} necessary to perform Bayesian inference over the model parameters.
As before, $\forall
s\in S, \forall \alpha \in \Act(s), \forall \param\in\Param : \sum_{s'\in S}\Tp_\param(s,\alpha, s') = 1$.

\subsection{Strategies}
A strategy for an MDP resolves nondeterminism by choosing an action in each state of the model. 
In our work
experiment design amounts to synthesising a strategy for an MDP, i.e., a sequence of actions, 
under which we generate data from the system. 
We focus on \textit{deterministic} \emph{memoryless} strategies in this paper, i.e., strategies that always pick the same action in any given state, independent of the history of states already visited. Future work will extend to both memory-dependent and randomised strategies. 
%and first consider memory-dependent ones, 
%which choose actions based on the history of their execution so far.

%\begin{definition}
%\label{def:scheduler}
%A deterministic \textit{memory-dependent strategy} for an MDP $\M$ is a function %$\strategy:S^{+}\rightarrow Act$, 
%such that for every $\omega=s_{0}s_{1}s_{2}...s_{n}\in S^{+}$,
%$\strategy(\omega)\in Act(s_{n})$, 
%where $S^{+}$ is a finite sequence of states ending in state $s_{n}$. 
%\end{definition}

\begin{definition}
A deterministic memoryless strategy for an MDP $M$ is a function $\strategy: S \rightarrow \Act \,\,s.t. \,\, \strategy(s)\in \Act(s) \,\,\forall_{s \in \sys}$.
\end{definition}
%\begin{definition}
%\label{def:schedulermemless}
%A strategy $\strategy$ is called \textit{memoryless} if and only if $\strategy(\omega)$ depends only on the last state in $\omega$. 
%That is, if and only if any two finite sequences of states, $\omega, \omega'$, 
%ending in the same $s_n$, conform to
%$\strategy(\omega) = \strategy(\omega')$. 
%Hence, $\strategy: S \rightarrow Act$. 
%%$\omega'=q_{0}q_{1}q_{2}...q_{n} \ale{\in S^{+}}$ and $\omega=s_{0}s_{1}s_{2}...s_{n}\in S^{+}$ with $s_{n}=q_{n}$,
%%we have that $\strategy(\omega)=\strategy(\omega')$; that is $\strategy:S \rightarrow Act$. 
%\end{definition} 

\subsection{Properties -- Probabilistic Computational Tree Logic}

We consider system specifications (aka properties) given in a fragment of Probabilistic Computational Tree Logic (PCTL)~\cite{Baier2008}. 
Since we use PRISM~\cite{DBLP:conf/cav/KwiatkowskaNP11} for parameter synthesis, 
we consider 
%only the set of properties supported by this synthesis tool: 
\textit{non-nested}, \textit{unbounded-time ``until''} properties expressed in PCTL. 
%Formulations in alternative logics such as Probabilistic Linear Temporal Logic, 
%as well as extensions to other formulae such as those expressing \textit{bounded-time} properties, 
%are possible but are depennot further discussed.
% 
\begin{defn}
\label{def:pctlsyn} 
Let a discrete-time MDP be given.  
Let $\prop$ be a formula interpreted over states $s\in S$, 
and $\pathprop$ be a formula interpreted on paths of the MDP.
Also, let $\bowtie\,\in\{< ,\,\leq, \,\geq ,\,>\}$, $n\in\mathbb{N}$, $p\in [0,1]$, $c\in AP$. 
The Syntax of the PCTL fragment we consider is given by: 
\[\prop:= \textrm{\normalfont true}\mid c\mid \prop \wedge \prop \mid \neg\prop \mid \Props_{\bowtie
p}(\pathprop),\hspace{1cm}\pathprop:= \bigcirc\,\prop\mid\prop\,\,
\mathcal{U}\,\,\prop.\]
\end{defn}
%\noindent A property (PCTL formula $\prop$) that is satisfied by an MDP $\M$ is
%written as: $\M\models \prop$.   

\begin{defn}
\label{def:pctluntilsat} 
Consider a PCTL formula $\prop:= \Props_{\bowtie p}(\prop_1\,\,\mathcal{U}\,\,\prop_2)$. 
Let $\mathbb{P}^{\strategy}_{\M}(s,\pathprop)$ denote the probability associated to the paths 
of an MDP $\M$ starting from $s\in S$ satisfying the path formula $\pathprop$ under the strategy $\strategy$. 
Let $\mathfrak{A}(\M)$ denote all deterministic memoryless strategies for $\M$. 
The satisfaction of the formula $\prop$ by $M$ is given by: 
\[
\M\models\Props_{\bowtie p}(\prop_1\,\,\mathcal{U}\,\,\prop_2) \Longleftrightarrow \forall s\in S, \iota_{init}(s)>0:  
\min_{\strategy\in \mathfrak{A}(\M)}\mathbb{P}^{\strategy}_{\M}(s,\prop_1\,\,\mathcal{U}\,\,\prop_2) \bowtie p.  
\]
\end{defn}

\noindent We introduce the \textit{feasible set} of parameters, denoted $\Param_{\prop}$, which is the set of parameter evaluations for which the property is satisfied. 

\begin{defn}
\label{def:feasible_set}
Let $\Mp$ be an induced MDP of the pMDP $\Mset$, 
indexed by parameter vector $\param\in \Param$.   
Let $\prop$ be a formula in PCTL. The feasible set $\Param_\prop$ is defined as: 
$\param \in \Param_\prop\Longleftrightarrow \M(\param) \models \prop$.
\end{defn}

\noindent We use $\mathbb{P}(A)$ to denote the probability of an event
 $A$, $p(\cdot)$ to represent probability density functions and
 $\Props_{\bowtie p}(\cdot)$ for the probabilistic operator in PCTL.
%\ale{
%\begin{remark}
%In this work we let $\mathbb{P}(\cdot)$ denote a probability measure over the MDP, 
%$p(\cdot)$ a probability density function (say, over the domain of a parameter), 
%whereas $\Props_\cdot(\cdot)$ characterises a probabilistic operator in PCTL.  \qed
%\end{remark}
%}

%%%%%%%%%%%%%%%%%%%%%%%%%%%%%%%%%%%%%%%%%%%%
%Section 3: Overview of our Approach
%%%%%%%%%%%%%%%%%%%%%%%%%%%%%%%%%%%%%%%%%%%%
% !TEX root = MDPpaper.tex

\section{Overview of the Method}
\label{sec:overview}

\begin{figure}[!htb]
    \centering
\begin{tikzpicture}[->,>=stealth' ,   state/.style={
           rectangle,
           rounded corners,
           draw=black, very thick,
           minimum height=2em,
           inner sep=2pt,
           text centered,
           }]
 \node[state, text width=3cm] (SYNTH) 
 {\textbf{1: }Parameter \\synthesis};

  \node[state,   rounded corners=0pt,  draw=black,   thick,double, rectangle, above of = SYNTH, xshift=-0.75cm, node distance=1.7cm] (PROP) 
 { Property $\prop$ };
 
 \node[state,      rounded corners=0pt, xshift=0.3cm,   draw=black,   thick,double, rectangle,right of=PROP, node distance=2cm](MODEL)
 {Model pMDP};
  
 \node[state, text width=3cm, right of=SYNTH, node distance=4cm,
  anchor=center] (STRAT) 
 {\textbf{2a: }Strategy\\ synthesis};

 \node[state, text width=3cm, above right of=STRAT,  node distance=2.5cm, xshift=0.5cm ,
  anchor=center] (SYS)
 {\textbf{2b: }Generate data from system};

 \node[state,text width=3cm,below right of=SYS,node distance=2.5cm,xshift=0.5cm ,
  anchor=center] (INF) 
 {\textbf{2c: }Bayesian inference over parameters};

 \node[state,below  of=STRAT,node distance=1.5cm,text width=3cm, anchor=center] (CONF) 
 {\textbf{3: }Confidence computation};
 \node (connectConf) at ([yshift=.166cm]CONF.west) {};
 \node (connectConftwo) at ([yshift=-.166cm]CONF.west) {};

 \node [          draw=black,   thick,double, rectangle, below of =SYNTH, node distance =1.66cm,text width=3cm,   minimum height=2em,
           inner sep=2pt,           text centered,
 ] (RESULT){$\conf = \mathbb{P}(\sys \models \prop)$};

 % draw the paths and and print some Text below/above the graph
 \path 
 (SYNTH) edge node [above]{$\Param_\prop$} (STRAT)
 (STRAT) edge node [sloped, above]{$\strategy$}(SYS)
 (SYS)   edge node [right,xshift=0.2cm]{$\Dt$}(INF)
 (INF)   edge node [sloped, above] {$p(\param {\mid} \Dt)$} (STRAT);
\draw  (INF)  |- node [sloped, right]{$p(\param {\mid} \Dt)$} (CONF.east);
\draw  (SYNTH.east)  -|(+2cm, 0cm)|- %node [above]{$\Param_\prop$}
(connectConf.center);
\draw  (connectConftwo.center)  |- (RESULT);
\draw[<-]   (SYNTH.north)+ (-.1cm,0cm)--([yshift=.5cm, xshift=-.1cm]SYNTH.north)-| (PROP.south);
\draw[<-]  (SYNTH.north) +(+.1cm,0cm) --([yshift=.5cm, xshift=+.1cm]SYNTH.north)-| (MODEL.south);
\end{tikzpicture}
    \caption{Overview of the verification procedure.}
    \label{fig:blockdiag}
  \end{figure}
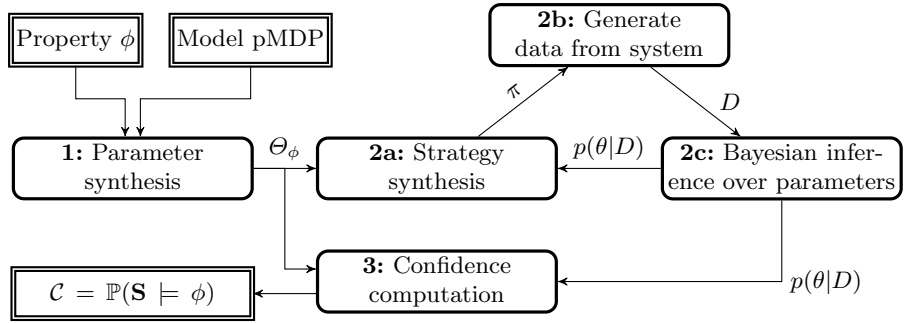

Our method is made up of three distinct phases, 
as shown in Fig.~\ref{fig:blockdiag}. 
\begin{enumerate}

\item 
We use a parameter synthesis tool to determine a set of feasible parameters for which the property is satisfied by the system, 
based on the given parametric Markov decision process, see Section~\ref{sec:param_synth}.
\item 
	\begin{enumerate}
	\item
	We synthesise a strategy for collecting data, 
	based on the feasible set and the prior distribution over the parameters, 
	see Section~\ref{sec:strat_synth}. 
	\item 
	We collect data from the underlying system using the synthesised strategy, 
	see Section~\ref{sec:data}. 
	\item 
	We use Bayesian inference to infer a distribution over the likely values of the parameters, 
	based on the collected data, 
	and update the respective prior distributions with the new information, see Section~\ref{sec:inf}. 
	If we can sequentially collect more data, loop back to step 2 (a).
	\end{enumerate}
\item 
We compute the confidence that the system satisfies the property, 
based on the data collected, see Section~\ref{sec:conf}. 
\end{enumerate}
Updating the posterior distributions for parametric Markov decision processes with linear relationships between
the parameters requires special treatment, as detailed in Section~\ref{sec:linear_param}.

%\ale{[it seems like we're bound to loop back to step 2 from 4 only once, whereas we could leave the option to break down the procedure in more loops?]} 

%%%%%%%%%%%%%%%%%%%%%%%%%%%%%%%%%%%%%%%%%%%%
%Section 4: Bayesian inference and parameter synthesis in MDPs
%%%%%%%%%%%%%%%%%%%%%%%%%%%%%%%%%%%%%%%%%%%%
\vspace{-1cm}
\subsection{Parameter Synthesis}
\label{sec:param_synth}

The first phase of the method uses parameter synthesis to find the feasible set of parameters, 
namely parameter evaluations corresponding to models of the considered  pMDP that satisfy the given PCTL property. 
This step leads to the set of parameters $\Theta_\phi = \{\theta\in\Theta : \M(\theta)\models\phi\}$.

The output of the parameter synthesis procedure is a mapping from hyper-rectangles 
(which are subsets of parameter evaluations) to truth values, namely ``true'' 
if the property is satisfied in the hyper-rectangle and ``false'' otherwise.

\paragraph{Implementation:} We use PRISM~\cite{DBLP:conf/cav/KwiatkowskaNP11} for parameter synthesis: 
the tool computes a rational function of the parameters, 
which expresses the result obtained from model checking the PCTL property on the parameterised model. Our approach can also make use of Storm~\cite{Quatmann2016}, 
which shows potential
to be scalable to much larger systems. 
Storm \textit{lifts} a parametric Markov decision process to a parameter-free Stochastic Game (SG) between two players, and solves the resulting SG via standard value iteration. 

\section{Bayesian Inference in Parametric Markov Decision Processes}
\label{sec:inf}

In this work, we collect data from the underlying system and
use Bayesian learning to infer a probability 
distribution over parameters of the pMDP model based on the collected data. 
Bayesian inference maintains a
probability distribution over these parameters and updates the distribution by employing Bayes' rule as more
observations are gathered~\cite{DBLP:journals/jmlr/RossPCK11}. 
%This procedure is an extension to Markov Decision Processes of~\cite{DBLP:conf/qest/PolgreenWHA16}, 
%which applies the same principles to Markov chains. 
An initial prior distribution $p(\param)$ 
%must be specified or 
is assumed.
%which is then updated using Bayes rule after a set of traces is collected from the system. 

\subsubsection{Data.}
\label{sec:data} 

We collect \textit{finite traces} from the underlying system, in the form of a sequence of visited
states and actions. We use $\Dt$ to denote a set of finite traces. 
We split the data into transition counts: 
$\Dt_{s_k,\alpha_1, s_l}$ denotes the number of
times the transition from $s_k$ to $s_l$ under action $\alpha_1$ appears within the data set $\Dt$. Each transition count is the outcome of an independent trial in a multinomial distribution\footnote{A multinomial distribution is defined by its density function $f(\cdot \mid p, N) \propto
\prod_{i=1}^{k}p_i^{n_i}$, for $n_i \in \{0,1,...,N\}$ and such that $\sum_{i=1}^{k}n_i = N$, where $N \in \mathbb{N}$
is a parameter and $p$ is a discrete distribution over $k$ outcomes.} 
%
% \ale{[not introduced before]}, 
with event probabilities given by the transition probabilities. 

Assume for now that the transitions are parameterised either with constants or with single parameters of the form $\param_i$ or $1-\param_i$.
We can group transition counts for identically parameterised transitions. We shall denote by $\Dt_{\param_j}$ the transition counts for all transitions with probability given by $\param_j$. 
%\ale{[of course, some states are linearly parameterised \ldots]}

We wish to obtain posterior distributions for each parameter via {\it marginal distributions} 
(which, in this case, are binomial distributions), 
by applying \textit{parameter-tying}~\cite{DBLP:conf/icml/PoupartVHR06} techniques. 
We thus obtain a number of transition counts for $1 - \param_j$ as the sum of all transitions
not parameterised with $\param_j$, 
under an action that has a transition parameterised with $\param_j$, 
and denote it by $\Dt_{\neg\param_j}$. %Formally, 
Hence $\Dt_{\param_j}$ and $\Dt_{\neg\param_j}$ are calculated as:

\noindent\scalebox{0.97}{\parbox{1\textwidth}{%
\begin{align*}
\Dt_{\param_j} &=  \!\!\!\!\!\!\!\!\sum_{s_i \in S, s_l\in S, \alpha_k\in Act}  \!\!\!\!\Dt_{s_i, \alpha_k,s_l}\textmd{ for�} \, \Tp(s_i,\alpha_k, s_l)=\param_j, \mbox{ and}\\
\Dt_{\neg\param_j} &= \!\!\!\! \!\!\!\!\sum_{s_i \in S, s_l\in S, \alpha_k\in Act} \!\!\!\! \Dt_{s_i, \alpha_k,s_l}\textmd{ for }\Tp(s_i,\alpha_k, s_l)\neq\param_j 
\wedge \exists s_m\in S: \, \Tp(s_i,\alpha_k, s_m)=\param_j.
\end{align*}
}}

\noindent Let $\Dt_{\param_j, \neg\param_j}$ denote the pair $(\Dt_{\param_j}, \Dt_{\neg\param_j})$.
For parameterisations where the transition probabilities are expressed as linear functions of parameters, 
we obtain $\Dt_{\param_j, \neg\param_j}$ by the same procedure that~\cite{DBLP:conf/qest/PolgreenWHA16} uses, 
in which transition probabilities expressed as multinomial distributions. 
Further details can be found in Section~\ref{sec:linear_param}.

\subsubsection{Bayesian Inference with Data.}

Consider a parametric Markov decision process $\Mset = (S$, $\Act$, $\Tp_\param$, $\iota_{init}$, $\AP$, $L$, $\Param)$ 
with $\Param\subseteq  [0,1]^{n}$. 
Suppose that we have obtained $\Dt_{\param_j}$ and $\Dt_{\neg\param_j}$ for all $\param_j \in \param$, 
and that we have assumed non-informative, 
uniform prior distributions for all parameters $\param_j \in \param$, denoted by $p(\param_j)$.
The posterior density $p(\param_j \mid \Dt)$ is given by Bayes' rule: 
\begin{align*} 
p(\param_j\mid \Dt) &=
\frac{\Prob(\Dt\mid \param_j) p(\param_j)}{\Prob(\Dt)} 
=\frac{p(\param_j) \param_j^{\Dt_{\param_j}} (1-\param_j)^{\Dt_{\neg\param_j}}}
{\Prob(\Dt_{\param_j, \neg\param_j})}.
\end{align*} 

%\ale{[why do we have different notations for probability?]}\sof{Cause one of them is a probability density $p$ (a function whose integral over an event set assigns the probability measure of that event), the other one should be the probability measure. For the latter I have found $\mathbf P$ (sec.2.2.), $\mathbb P$ and $P$ (fig.1.) in the draft.... Better fix this! }

%As such, 
%$p(\param_j \mid \Dt)$ depends on $\Dt_{\param_j}$ and $\Dt_{\neg\param_j}$, i.e., the number
%of times a transition parameterised with $\param_j$ or $1-\param_j$ features in the data set
%$\Dt$.
%The likelihood function %$\param_j^{\Dt_{\param_j}}(1-\param_j)^{\Dt_{\neg\param_j}}$ 
%is in
%the form of a binomial distribution. 
A standard approach~\cite{DBLP:conf/nips/FriedmanS98,DBLP:journals/csda/PasanisiFB12,DBLP:journals/jmlr/RossPCK11}
is to consider the prior to be a Dirichlet distribution. 
The posterior distribution is then updated by adding the event counts to the hyperparameters of the prior.
The Dirichlet prior distribution for the pair $(\param_{j},1 - \param_{j})$ is denoted as
$\Dir(\param_j \mid \mu^{\param_j})$ 
with hyperparameters $\mu^{\param_j} = (\mu_1^{\param_j}, \mu_2^{\param_j})$. Thus, the updated posterior
distribution for the parameter $\param_j$ is given as: $\param_j\sim p(\param_j \mid \Dt) =
\Dir(\param_j \mid \Dt_{\param_j,\neg\param_j} + \mu^{\param_j})$.%, where 
%
%\begin{align}
%\label{eq:dirichderi6} 
%p(\param_j \mid \Dt) 
%\propto p(\param_j)\param^{\Dt_{\param_j}} (1-\param_j)^{\Dt_{\neg\param_j}}
%\propto \param^{\mu_1^{\param_j} -1 + \Dt_{\param_j}} (1-\param_j)^{\mu_2^{\param_j} -1 + \Dt_{\neg\param_j}}. 
%\end{align}
%The normalisation constant of the obtained Dirichlet distribution for $p(\param_j \mid \Dt)$ is
%$B(\mu^{\param_j} + \Dt_{\param_j,\neg\param_j})$ %=  \Gamma(\mu_1^{\param_j} + \Dt_{\param_j}) \Gamma(\mu_2^{\param_j} +
%\Dt_{\neg\param_j})/\Gamma (\mu_1^{\param_j} + \Dt_{\param_j} + \mu_2^{\param_j} + \Dt_{\neg\param_j})$.  
%We can hence update the posterior probability distribution $p(\param_j \mid
%\Dt)$ by updating the parameters of a Dirichlet distribution as data is gathered. 

The posterior distribution
for the entire parameter vector $\param$, given by $p(\param \mid \Dt)$ is equal to the product of
the posterior distributions for all $\param_i \in \param$. 
This holds due to the independence of each $\param_i$ over independent state-action
pairs in the pMDP. %This notion of independence is also responsible for the independent
%prior distributions and likelihood functions for each $\param_i$. 
Note that, if we have a linearly parameterised MDP, we obtain some of the transition counts 
in the form of multinomial distributions. We hence obtain realisations of the posterior by a sampling procedure
from~\cite{DBLP:conf/qest/PolgreenWHA16} as explained in Section~\ref{sec:linear_param}.

 %\ale{sub-vector - component?}. 
%In conclusion, we have that: 
%\begin{equation*}
%p(\param \mid \Dt) = 
%\prod_{\param_i \in \param} \Beta(\param_i \mid \Dt_{\param_i, \neg\param_i} + \mu^{\param_i})
%\end{equation*}

%%%%%%%%%%%%%%%
%%%%%%%%%%%%%%%%%%%%%%%%%%%%%%
%%%%%%%%%%%%%%%%%%%%%%%%%%%%%%%%%%%%%%%%%%%%%
%\section{Appendix}

\subsubsection{Extension: obtaining $\Dt_{\param_j,\neg\param_j}$ for Linear Parameterisations.}
\label{sec:linear_param}

%\ale{[note on :=, to be discussed ]}

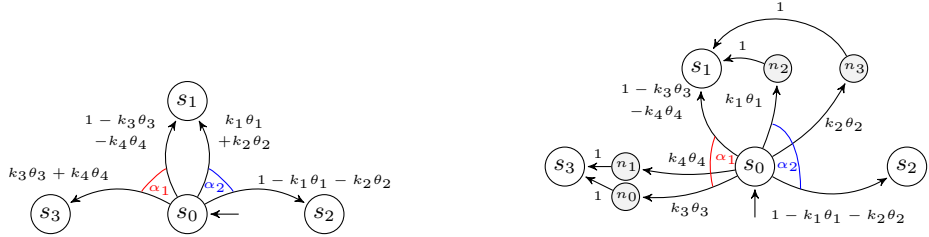
\begin{figure}[h]
    \centering
    \vskip -0.5cm
\begin{tikzpicture}[>=stealth',initial text={},shorten >=1pt,auto,node distance=2cm, scale = 1, transform shape]
\baselineskip=8pt
\node[initial right,blob] (S0) {$\!s_0\!$};
\node[blob](S1) [above of = S0, node distance=1.5cm]{$s_1$};
\node[blob](S2) [right of = S0, node distance=1.8cm]{$s_2$};
\node[blob](S3) [left of = S0, node distance=1.8cm]{$s_3$};

\path[->] (S0) edge [bend left] node[bend left, above left,align=center] {\tiny$1-k_3\param_3$\\ \tiny$- k_4\param_4$} (S1);
\path[->] (S0) edge [bend right] node[ above right,align=center] {\tiny$k_1\param_1$\\ \tiny$+ k_2\param_2$} (S1);
\path[->] (S0) edge [bend right] node[above left] {\tiny$k_3\param_3 + k_4\param_4$} (S3);
\path[->] (S0) edge [bend left] node[right,yshift=.1cm,xshift=-.1cm,align=center] {\tiny$1-k_1\param_1  -k_2\param_2$} (S2);

\draw pic[draw, red, angle radius=6mm,"\tiny $\alpha_1$"{xshift=0.105cm,yshift=-0.15cm},angle eccentricity=1.1, shorten >=0.3cm, shorten <=0.28cm] {angle = S1--S0--S3};
\draw pic[draw, blue, angle radius=6mm,"\tiny $\alpha_2$"{xshift=-0.15cm,yshift=-0.15cm},angle eccentricity=1.2, shorten >=0.28cm, shorten <=0.3cm] {angle = S2--S0--S1};
\end{tikzpicture}\hfill
\begin{tikzpicture}[>=stealth',initial text={},shorten >=1pt,auto,node distance=2cm, scale = 1, transform shape]
\node[initial below,blob] (T0) [ node distance =3cm]{$\!s_0\!$};

\node[blob](T2) [right of = T0]{$s_2$};
\node[newblob](N0)[left of =T0, node distance=1.7cm, yshift=-0.4cm]{\tiny$n_0$};
\node[newblob](N1)[left of =T0, node distance=1.7cm, yshift=0cm]{\tiny$n_1$};
\node[blob](T3) [left of = T0,node distance=2.5cm]{$s_3$};
\node[newblob](N2)[above of =T0, node distance=1.7cm, yshift=-0.4cm,xshift=.3cm]{\tiny$n_2$};
\node[newblob](N3)[right of=N2, node distance=1cm]{\tiny$n_3$};
\node[blob](T1) [left of = N2, node distance=1cm]{$s_1$};

\path[->] (T0) edge [bend left=15]  node[below, align=center] {\tiny$k_3\param_3$} (N0);
\path[->] (T0) edge [bend left=10]  node[above, align=center] {\tiny$k_4\param_4$} (N1);
\path[->] (N0) edge node[below, align=center] {\tiny$1$} (T3);
\path[->] (N1) edge node[above, align=center] {\tiny$1$} (T3);
\path[->] (T0) edge [bend left] node[above left, xshift=-.1cm,align=center,yshift=0.0cm] {\tiny$1-k_3\param_3$\\[-.3em] \tiny$- k_4\param_4$} (T1);
\path[->] (T0) edge [bend right=10] node [above left] {\tiny$k_1\param_1$}(N2);
\path[->] (T0) edge [bend right=15] node [ right] {\tiny$k_2\param_2$}(N3);
\path[->] (N2) edge  [bend right=20] node [above] {\tiny$1$}(T1);
\path[->] (N3) edge  [bend right=60] node [above] {\tiny$1$}(T1);
\path[->] (T0) edge [bend right] node[below,yshift=-.1cm, xshift=.1cm,align=center] {\tiny$1-k_1\param_1 - k_2\param_2$} (T2);

\draw pic[draw, red,angle radius=6mm,"\tiny $\alpha_1$"{xshift=-0.05cm,yshift=-0.05cm},shorten <=0.27cm, shorten >=-.13cm] {angle = T1--T0--N0};
\draw pic[draw, blue, angle radius=6mm,"\tiny $\alpha_2$"{xshift=-0.13cm,yshift=-0.45cm},angle eccentricity=1.2, shorten >=+0.1cm, shorten <=-0.3cm] {angle = T2--T0--N2};

\end{tikzpicture}
	    \caption{Transformation of a linearly-parameterised MDP: \textit{transition splitting}.}
    \label{fig:trans_splitting}
\end{figure}

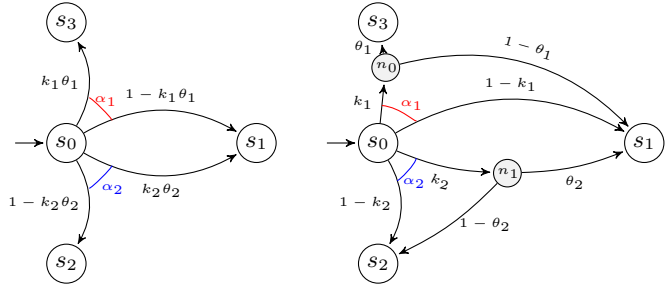
\begin{figure}[!htb]
\centering
    \vskip -0.3cm
\begin{tikzpicture}[>=stealth',initial text={},shorten >=1pt,auto,node distance=1.6cm, scale = 1, transform shape]
\baselineskip=8pt
\node[initial,blob] (S0) {$\!s_0\!$};
\node[blob](S1) [right of = S0, node distance=2.5cm]{$s_1$};
\node[blob](S2) [below of = S0]{$s_2$};
\node[blob](S3) [above of = S0]{$s_3$};

\path[->] (S0) edge [bend left] node[bend left, above, align=center] {\tiny$1-k_1\param_1$} (S1);
\path[->] (S0) edge [bend right] node[sloped, below, align=center] {\tiny$ k_2\param_2$} (S1);
\path[->] (S0) edge [bend right] node[left] {\tiny$k_1\param_1$} (S3);
\path[->] (S0) edge [bend left] node[left, align=center] {\tiny$1-k_2\param_2$} (S2);

\draw pic[draw, red, angle radius=6mm,"\tiny $\alpha_1$"{xshift=0.01cm,yshift=0.05cm},angle eccentricity=1.2, shorten >=0.3cm, shorten <=0.3cm] {angle = S1--S0--S3};
\draw pic[draw, blue, angle radius=6mm,"\tiny $\alpha_2$"{xshift=0.1cm,yshift=-0.1cm},angle eccentricity=1.2, shorten >=0.3cm, shorten <=0.3cm] {angle = S2--S0--S1};

\node[initial,blob] (T0) [right of=S1]{$\!s_0\!$};
\node[blob](T1) [right of = T0, node distance=3.5cm]{$s_1$};
\node[blob](T2) [below of = T0]{$s_2$};
\node[newblob](N0)[above of =T0, node distance=1cm, xshift=0.1cm]{\tiny$n_0$};
\node[blob](T3) [above of = T0]{$s_3$};
\node[newblob](N2)[right of =T0, node distance=1.7cm, yshift=-0.4cm]{\tiny$n_1$};

\path[->] (T0) edge node[left, align=center] {\tiny$k_1$} (N0);
\path[->] (N0) edge node[left, align=center] {\tiny$\param_1$} (T3);
\path[->] (N0) edge [bend left] node [sloped, above, align=center] {\tiny$1 - \param_1$} (T1);
\path[->] (T0) edge [bend left] node [sloped, above, align=center] {\tiny$1-k_1$} (T1);
\path[->] (T0) edge [bend right=10] node [below, sloped] {\tiny$k_2$}(N2);
\path[->] (N2) edge  [bend right=10] node [below] {\tiny$\param_2$}(T1);
\path[->] (N2) edge  [bend left=10] node [right] {\tiny$1 - \param_2$}(T2);
\path[->] (T0) edge [bend left] node[left, align=center] {\tiny$1-k_2$} (T2);

\draw pic[draw, red,angle radius=5mm,"\tiny $\alpha_1$"{xshift=0.0cm,yshift=0.1cm},angle eccentricity=1.2, shorten <=0.3cm, shorten >=-0.0cm] {angle = T1--T0--N0};
\draw pic[draw, blue, angle radius=5mm,"\tiny $\alpha_2$"{xshift=0.05cm,yshift=-0.11cm},angle eccentricity=1.2, shorten >=0.2cm, shorten <=0.25cm] {angle = T2--T0--T1};

\end{tikzpicture}
	    \caption{Transformation of a linearly-parameterised MDP: \textit{state splitting}.}
    \label{fig:state_splitting}
\end{figure}

%In this section, we explain how to obtain $\Dt_{\param_j, \neg\param_j}$ for linearly parameterised MDPs. 
Linearly parameterised MDPs, as stated in Definition \ref{def:pMDP}, have transition probabilities expressed using affine
functions of the form $g_l(\param)=k_0 + k_1\param_1+\ldots+k_n\param_n$. 
We apply two transformations, as introduced in~\cite{DBLP:conf/qest/PolgreenWHA16}, to the pMDPs.
The transformations result in an expanded model that contains only transition probabilities expressed as constants, 
or in the form of $\param_j$ or $1 - \param_j$, 
for any parameter $\param_j \in \param$. 
The expanded model allows us to derive \emph{distributions} for
$\Dt_{\param_j, \neg\param_j}$ for all component parameters $\param_j \in \param$, which we denote by $\Dt^*_{\param_j, \neg\param_j}$. We extend this notation and $\Dt^\ast_{s_k, \alpha_1, s_l}$ denotes the probability distribution of the transitions from $s_k$ to $s_l$ under action $\alpha_1$ in the expanded model, and $\Dt^\ast_{\param, \neg\param_j}$ denotes the distribution of $\D_{\param_j, \neg\param_j}$ in the expanded model.
We then present a procedure for performing Bayesian inference over these distributions. The two transformations are: 
\begin{itemize}
\item{\emph{Transition splitting}}, which expands a transition with probabilities expressed as
 $k_0 + k_1\param_1+\ldots+k_n\param_n$ into $n$ transitions with probabilities expressed as
  $k_0$, $k_1\param_1$,..., $k_n\param_n$, respectively. 
  This is illustrated in Fig.~\ref{fig:trans_splitting}.
\item{\emph{State splitting}}, which expands a transition with probabilities expressed as  $k_i\param_i$, into transitions with constant probabilities $k_i$
and $1 - k_i$, and transitions with single parameter probabilities $\param_i$ and $1 - \param_i$. 
This is illustrated in Fig.~\ref{fig:state_splitting}.
\end{itemize}
The expansions stated above are shown to be transitive and generally applicable to any linearly parameterised
Markov decision process. 
The new model, however, has transitions that did not feature in the original model, and
hence we are unaware of the corresponding exact transition counts. Therefore, we handle these new transition counts as 
%for the transitions we have introduced are now in the form of 
multinomial distributions over the transition probabilities.

As an example of transition splitting, consider Fig.~\ref{fig:trans_splitting} again. The expansion of the MDP introduces 4 new states,
$n_0, n_1, n_2, n_3$. The transition counts $\Dt^*_{s_0,\alpha_1, n_0}$ and
$\Dt^*_{s_0,\alpha_1, n_1}$ are unknown, but we know the total must be equal to $\Dt_{s_0, \alpha_1, s_3}$.
Hence, they follow the binomial distribution,  
$$
P(\Dt^*_{s_0,\alpha_1,n_0}=N) = {\Dt_{s_0,\alpha_1,s_3} \choose N}(\frac{k_3\param_3}{k_3\param_3 + k_4\param_4})^N
(\frac{k_4\param_4}{k_3\param_3 + k_4\param_4})^{\Dt_{s_0,\alpha_1,s_3}-N}. 
$$ 

\noindent The procedure for state splitting is similar; consider Fig.~\ref{fig:state_splitting}. The expansion
of the MDP introduces 2 new states. The transition counts $\Dt^*_{s_0,\alpha_2, n_1}$, $\Dt^*_{s_0,\alpha_2, s_2}$
and $\Dt^*_{n_1,s_2}$ are amongst the unknown transition counts. However, we know the counts for
$\Dt^*_{s_0,\alpha_2, s_2} + \Dt^*_{n_1,s_2}$ is equal to $\Dt_{s_0, \alpha_2, s_2}$ because the total counts into state $s_2$ must remain equal.
Hence we can specify the binomial distribution: 

\noindent\scalebox{0.97}{\parbox{1\textwidth}{%
$$
P(\Dt^*_{s_0,\alpha_2,s_2}=N) = {\Dt_{s_0,\alpha_2,s_2} \choose N}(\frac{1 - k_2}{1 - k_2 + 1 - \param_2})^N
(\frac{1 - \param_2}{1 - k_2 + 1 - \param_2})^{\Dt_{s_0,\alpha_2,s_2}-N}. 
$$ 
}}

We also know that the total counts into $s_1$ remain equal after the expansion and therefore $\Dt^*_{n_1,s_1} = \Dt_{s_0, \alpha_2, s_1} $. Using this method, we can find distributions that represent all unknown transition counts.
Note that these distributions depend on several $\param_i \in \param$:
we explain how we perform Bayesian inference over these distributions in the following section.

% !TEX root = MDPpaper.tex

\vspace{-0.5cm}
\subsubsection{Bayesian Inference with Distributions over Transition Counts.}
\label{sec:linear_param2}

We use $\mathcal{D}^*_{\param_j, \neg\param_j}$ to denote the set of all possible completions of $\Dt^*_{\param_j, \neg\param_j}$. 
We can apply Bayes' rule over the distributions over the transition counts as: 
\begin{equation*}
\pd{\param_j | \Dt_{\param_j, \neg\param_j}}= \sum_{\Dt^\ast_{\param_j, \neg\param_j}\in \mathcal{D^\ast}} 
\pd{\param_j |\Dt_{\param_j,\neg\param_j}^*}\Prob(\Dt_{\param_j, \neg\param_j}^*|\Dt). 
\end{equation*}
As mentioned before, completed data sets have a multinomial distribution dependent on the parameterisation, hence
the distribution of $\Dt^*_{\param_j, \neg\param_j}$ is given by $\Prob(\Dt^*_{\param_j, \neg\param_j})=\int_\Param\Prob(\Dt^*_{\param_j, \neg\param_j}|\param_j) \pd{\param_j}d\param_j$.
For a given $\Dt_{\param_j, \neg\param_j}$, the conditional distribution 
%$\Prob(\Dt^*_{\param_j, \neg\param_j}|\Dt)$ is given by 
$\Prob(\Dt^\ast_{\param_j, \neg\param_j}|\Dt_{\param_j, \neg\param_j})$ is 
$\Prob(\Dt^*_{\param_j, \neg\param_j})/\Prob(\Dt_{\param_j, \neg\param_j})$ with $\Dt_{\param_j, \neg\param_j}^* \in \mathcal{D}^*_{\param_j,\neg\param_j}$ and
$\Prob(\Dt)=\sum_{\mathcal{D}^\ast}\int_\Param\Prob(\Dt^*_{\param_j, \neg\param_j}|\param_j)\pd{\param_j} d\param_j$.

Realisations of the posterior $\pd{\param_j |\Dt_{\param_j,\neg\param_j}^*}$ can be obtained by sampling without computing the entire integral.
We generate a set of $N$ samples of $\Dt_{\param_j, \neg\param_j}^*$ by sampling 
from the distribution
$\Prob(\Dt^*_{\param_j, \neg\param_j} \mid \Dt)$ and then generate a sample of $\param_j$ from 
the distribution $\Prob(\param_j \mid \Dt^*_{\param_j, \neg\param_j})$ for each sample of $\Dt^*_{\param_j, \neg\param_j}$. 
These samples are then used directly to compute the confidence.

%\ale{check consistent use of Pr vs $\Prob$ vs p vs bold P (in pctl syntax)}

%%%%%%%%%%%%%%%%%%%%%%%%%%%%%%%%%%%%%%%%%%%%
%Section 5: Confidence calculation
%%%%%%%%%%%%%%%%%%%%%%%%%%%%%%%%%%%%%%%%%%%%
\section{Computation of Confidence}
\label{sec:conf}

%Confidence in a property can be viewed as a measure of uncertainty over a synthesised
%parameter set. Such uncertainty can be quantified by a probability distribution of random
%variables[cite bayesian probability calculus]. Formally, for a partly unknown dynamical system
%$\mathbf{S}$, the confidence in $\mathbf{S}\models\phi$ can be quantified via Bayesian inference as
%$\mathbb{P}(\mathbf{S}|D)=\int_{\Theta}f_{\phi}(\theta)p(\theta|D)d\theta$ for a given data set $D$
%where $\theta\in\Theta$ and $\phi$ is a property expressed in PCTL.

We determine a confidence, $\conf$, for the satisfaction of a PCTL
formula $\prop$ by
a system $\mathbf{S}$ of interest. We first presented this
procedure in previous work~\cite{DBLP:conf/qest/PolgreenWHA16}, and we
need no extension to this due to the external nondeterminism being factored out in the
Bayesian inference calculation given in the previous section. 
%The procedure requires as input a posterior
%distribution for an evaluation $\param$ over $\Param$ and 
%results from the parameter synthesis phase, i.e., the feasible set of
%parameters for $\mathbf{M}(\param) \models\prop$.

\begin{definition}\label{def:confidence}
Given a PCTL formula $\prop$ that has a binary satisfaction function, i.e., the property is either
satisfied or not, 
and posterior distributions $p(\param_i \mid \Dt)$ for all $\param_i \in \param$, as obtained in the previous section,
%a set of finite, complete traces (sample trajectories) $\Dt$ from the system $\textbf S$; 
% and an independent parameterisation (as obtained by the two aforementioned transformation procedures)
% \todo{\brian{``we dont explicitly state anything about transformations in this section"}\ep{Yes, that's because the transformations are done as part of bayesian inference, not the conf calc}}
the confidence in $\textbf S\models\prop$ can be quantified by Bayesian inference as 
\begin{equation}
\label{eq:confiref}\textstyle
\conf = \mathbb{P}(\textbf S \models\prop \mid \Dt) 
=\int_{\Param_{\prop}}\prod_{\param_i \in \param} 
p(\param_i\mid \Dt_{\param_i, \neg\param_i})d\param,  
\end{equation}
%\ale{[well, we have a different notation for probability than in figure 1 it seems]}\sof{[Also the formula is different...]}
%where $\Param_{\prop}$ denotes the feasible set of parameters corresponding to models verifying the
%property $\prop$ (as generated by PRISM). \brian{we talked about feasible set number of times before so do we need to restate its definition here?}
\end{definition}
The operation shown in Eq.~\eqref{eq:confiref} is equivalent to computing the confidence that each parameter is within its feasible set, 
and then taking the product of all the parameter confidence values. 
The integral of a Dirichlet distribution is hard to compute using analytical methods, 
and so we use Monte Carlo integration. 
This also allows integration with the calculation of the posterior distribution for pMDPs with linear parameterisations, 
where we have obtained the posterior distribution by means of sampling, as described in Section~\ref{sec:linear_param}.

 %therefore, we resolve to iterative or numerical methods as a general solution by
%using a simple Monte-Carlo approach as presented in Algorithm 2 of our previous
%work~\cite{DBLP:conf/qest/PolgreenWHA16}. This allows good integration with the calculation of the
%posterior distribution using the \textit{state splitting} transformation.

%Section 6: strategy choice
% !TEX root = MDPpaper.tex
\section{Online Experiment Design}
\label{sec:strat_synth}

The key contribution in this paper is the design of experiments to generate maximally useful data. 
We describe in the preceding sections how we use a limited amount of data efficiently to obtain a confidence that the system satisfies the property.
 In this section, we propose a method for
selecting the deterministic memoryless strategy that provides the most useful data to input into our confidence computation in Section~\ref{sec:conf}. This allows us to compute the most accurate confidence value for the finite
data set of limited size, i.e., the confidence should be high if the underlying system satisfies the property, 
and low if the underlying system does not satisfy the
property.

\subsection{Predicted Confidence}
\label{sec:ex_conf}

We predict the confidence after 
taking a transition from state $s$ under action $\alpha$.
%we  define the predicted value of the confidence to be an approximation of this and denote it as 
%$\Pred{\conf_{s, \alpha}}$.  
%More precisely, the 
We define the predicted confidence, $\Pred{\conf_{s, \alpha}}$, to be 
the confidence computed using the \textit{expected parameter counts}, 
after taking a single transition from $s$ under action $\alpha$: 
these are denoted by $\EXsa{\Dt_{\param_i, \neg\param_i}}$
for all $\param_i \in \param$. Formally, \begin{equation*}
\Pred{\conf_{s, \alpha}} =\int_{\Param_{\prop}}\prod_{\param_i \in \param} p(\param_i\mid \EXsa{\Dt_{\param_i, \neg\param_i}})d\param, 
\end{equation*}
where $p(\param_i\mid \EXsa{\Dt_{\param_i,\neg\param_i}})$ is the predicted posterior distribution obtained by updating the prior, $Dir(\param_i \mid \mu^{\param_i})$, with the expected parameter counts, 
i.e., 
$\Dir(\param_i \mid \mu^{\param_i} + \EXsa{\Dt_{\param_i,\neg\param_i}})$.
% $\Dir(\mu_1^{\param_i} + \EX{\Dt_{\param_i}}, \mu_1^{\param_i} + \EX{\Dt_{\neg \param_i}})$.

We first compute the \textit{expected transition counts} for the state-action pair, $\EXsa{\Dt_{s, \alpha}}$, 
from which we extract the \textit{expected parameter counts}
using the method in Section~\ref{sec:data}.
Consider a state $s$ with an action $\alpha$, and
two transitions with probabilities 
$\T_\param(s, \alpha, s') = g_l(\param) = k_0 + k_1\param_1 + ... + k_n\param_n$, 
and $T_\param(s, \alpha,s) = 1 - g_l(\param)$. The expected transition counts 
are given by a multinomial distribution over the outgoing transitions under that action,
with event probabilities equal to the \textit{expected transition probabilities}.
Note that prior distribution for any parameter $\param_i \in \param$ is 
$Dir(\param_i \mid \mu^{\param_i})$.
%$\Dir(\mu_1^{\param_i}, \mu_2^{\param_i})$ \ale{[is this notation for Dir distributions aligned with that used before?]}. 
To compute the expected transition probabilities, we require the expected values of the 
parameters, given by $\EX{\param_i} = \frac{\mu_1^{\param_i}}{\mu_1^{\param_i}+\mu_2^{\param_i}}$ for all
$\param_i \in \param$. 
The expected value of the transition probabilities are then given by evaluating $g_l(\EX{\param})$
and $1 - g_l(\EX{\param})$. 
Hence the expected transition counts $\EXsa{\Dt_{s, \alpha, s'}}$ and $\EXsa{\Dt_{s, \alpha, s}}$, 
are equal to the expected transition probabilities for $\T_\param(s, \alpha, s')$ and $\T_\param(s, \alpha, s)$. 
Consider only the transition parameterised with $\T_\param(s, \alpha, s') = g_l(\param)$: 
\begin{align*} 
\EXsa{\Dt_{s,\alpha,s'}} &= \EX{\T(s,\alpha,s'}) = g_l(\EX{\param}) \\
				&= k_0 + k_1 \EX{\param_1} + ... + k_n \EX{\param_n} 
				  = k_0 + \sum_{i=1:n} k_i \frac{\mu_1^{\param_i}}{\mu_1^{\param_i}+\mu_2^{\param_i}}. 
\end{align*} 

\noindent We can extract the parameter counts as described in Section \ref{sec:data}, 
to obtain $ \EXsa{\Dt_{\param_i, \neg\param_i}}$.  

\subsection{Optimisation of Predicted Confidence Gain}

The underlying system either satisfies or does not satisfy the given property, 
so we wish to minimise the difference between our confidence value and the closest among $0$ or $1$, or
to maximise the difference between a confidence of $0.5$ and our confidence, 
i.e., the maximum absolute value of $0.5 - \conf$.
We can therefore define a predicted confidence gain for a state-action pair $(s,\alpha)$, 
denoted by $\Gain_{s, \alpha}$, 
as the maximisation of this difference, i.e., the biggest step towards either $0$ or $1$.
\vspace{-0.1cm}
\begin{equation*}					
\Gain_{s,\alpha} =  |0.5 -  \Pred{\conf_{s,\alpha}}| - | 0.5  - \conf| 
\end{equation*}
\vspace{-0.1cm}
For a finite trace of length $N$,  
we can calculate the optimal predicted confidence gain for state $s$ and discrete time step $t$, denoted by $x_s^t$, as 
\begin{align*}
x_s^t = \begin{cases}
			\max_{\alpha \in Act(s)} (\Gain_{s,\alpha} + \sum(\Tp(s,\alpha, s'). \, x_{s'}^{t+1})) & \quad \text{if } 0 < t < N \\
			0 & \quad \text{if } t \geq N. 
			\end{cases}
\end{align*}
It is important to note that the confidence gain is not a static quantity, because $\Gain_{s, \alpha}$
depends on the distribution over the relevant component parameters of $\param$ at time $t$.  

\subsection{Optimal Confidence Gain: Experiment Design via Strategy Synthesis}

Due to memory dependency of the confidence gain, 
computing an optimal strategy is intractable, 
and cannot be solved via conventional dynamic programming methods~\cite{DBLP:conf/uai/GrettonPT03}. 
However, we put forward a few alternatives.
\paragraph{Explicitly evaluated memoryless strategies.}
The conventional way of solving a MDP with non-Markovian rewards
is to translate the model into an equivalent MDP with Markovian rewards, whose states 
result from augmenting those of the original
model with extra information capturing enough history to make the reward Markovian. 
This is in general computationally expensive~\cite{DBLP:conf/uai/GrettonPT03}. 
Given that we will be performing strategy synthesis repeatedly in our method 
(i.e., once each time a new batch of data is sequentially gathered), 
we compromise and use a straightforward selection method to find the best memoryless strategy. 
This reduces the number of possible strategies and allows us to consider each possible strategy individually. 
We simplify the calculations in Section~\ref{sec:ex_conf} to compute the expected transition counts
for a full trace of length $N$, and then compute the predicted confidence gain for the entire memoryless strategy. This method works well for small trace lengths, 
however computing the expected transition counts for a full trace of length $N$ amounts to performing a matrix multiplication $N$ times, 
so this can be time consuming for large $N$. 

\paragraph{Alternative off-line method.}
An alternative approach would be to disregard the memory dependency of the confidence gain. 
This corresponds to an off-line approach:  
we compute a strategy on the model frozen at the time we start generating traces, 
assuming that the prior distributions remains unchanged over the trace horizon $N$. We assign confidence gains to state-action pairs and treat them as static rewards. 
This allows us to use classical dynamic programming to find the best memoryless strategy, 
which would require introducing a discount factor on the rewards, 
to avoid infinite returns inside \textit{strongly-connected components}. This method may be faster for long trace lengths than explicitly evaluating possible strategies, as done previously; 
however, the selected strategy may not be the best memoryless strategy when the trace lengths are large, 
and specifically when the prior distributions, which are assumed to remain unchanged, actually change significantly over time as the trace length is being reached. 
%\ale{[unclear: why not, because our problem is over a finite horizon? because of the discount factor? pls clarify]}. 

\begin{figure}[h]
\centering

\begin{tikzpicture}[>=stealth',initial text={},shorten >=1pt,auto,node distance=1.5cm, scale = 1, transform shape]
\baselineskip=8pt
\node[initial below,blob] (S0) {$\!s_0\!$};
\node[blob](S1) [above of = S0, node distance=1.5cm]{$s_1$};
\node[blob](S2) [left of = S0, node distance=2.2cm]{$s_2$};
\node[blob](S3) [right of = S1]{$s_3$};
\node[blob](S4) [right of = S0]{$s_4$};
\node[blob](S5) [right of = S4]{$s_5$};
\node[blob](S6) [right of = S3]{$s_6$};

\path[->] (S0) edge  node {\tiny$\param_1 + \param_2$} (S1);
\path[->] (S1) edge [loop left] node {\tiny$1$} (S1);
\path[->] (S2) edge [loop left] node {\tiny$1$} (S2);
\path[->] (S0) edge  node[above, align=center] {\tiny$1-\param_1 -\param_2$} (S2);
\path[->] (S0) edge node [sloped, align=center, above]{{\color{red}\tiny$\alpha_3$} \tiny$1$ }(S4);
\path[->] (S0) edge node [sloped, align=center, above]{{\color{red}\tiny$\alpha_2$} \tiny$1$ }(S3);
\path[->] (S4) edge [loop below] node  [ right] {\tiny$\param_1$}(S4);
\path[->] (S4) edge [bend left] node [sloped, above, align=center]{\tiny$1 - \param_1$}(S5);
\path[->] (S3) edge [loop below] node [swap] {\tiny$\param_1$}(S3);
\path[->] (S5) edge [loop below] node [ right] {\tiny$\param_1$}(S5);
\path[->] (S5) edge [bend left] node [sloped, above, align=center]{\tiny$1 - \param_1$}(S4);
\path[->] (S6) edge [loop below] node [swap] {\tiny$\param_2$}(S6);
\path[->] (S3) edge [bend left] node [sloped, above, align=center]{\tiny$1 - \param_1$}(S6);
\path[->] (S6) edge [bend left] node [sloped, above, align=center]{\tiny$1 - \param_2$}(S3);

\draw pic[draw, red, angle radius=4mm, "\tiny $\alpha_1$"{xshift=0.0cm,yshift=0.1cm},angle eccentricity=1.2, ] {angle = S1--S0--S2};

\end{tikzpicture}
\caption{Example pMDP where offline strategy synthesis may not be optimal}
\label{fig:example_missing_data}
\end{figure}
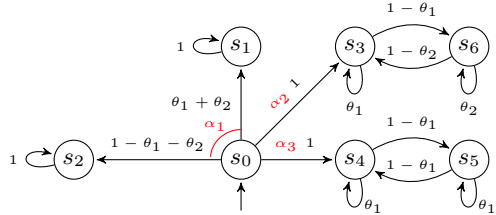

\paragraph{Comparison.}
Consider the small pMDP shown in Fig.~\ref{fig:example_missing_data},  
parameterised with $\param = (\param_1, \param_2)$, 
and the property $\Props_{\leq 0.5} (\text{true}\,\,\mathcal{U}\,\,s_1)$. 
Both parameters have the same prior distributions and both contribute equally 
to the feasible set. Intuitively, choosing action $\alpha_2$ or $\alpha_3$ is better than choosing action $\alpha_1$,
because any trace starting with $\alpha_1$ only contains one parameterised transition. However,
it is also intuitive that choosing $\alpha_2$ is better than $\alpha_3$
because any trace starting with $\alpha_3$ only gives us information about $\param_1$, 
whereas traces with $\alpha_2$ give us information about both parameters. 

The dynamic programming approach will pick nondeterministically between action $\alpha_3$ and $\alpha_2$
for the first trace, because the reward assigned to $(s_0, \alpha_3)$ is the same 
as the reward assigned to $(s_0, \alpha_2)$ as the initial priors and the feasible
sets are the same. The priors will not be updated until after the full trace is collected.
Our strategy synthesis approach calculates the expected updates for these priors, 
and will thus be able to detect a better strategy, 
which selects action $\alpha_2$. 

In our experimental evaluation, we use the explicitly evaluated memoryless strategy. Henceforth, the explicitly evaluated memoryless 
strategy will be referred to as the \textit{synthesised strategy}. 

%%%%%%%%%%%%%%%%%%%%%%%%%%%%%%%%%%%%%%%%%%%%
%Section 6: Experimental results
%%%%%%%%%%%%%%%%%%%%%%%%%%%%%%%%%%%%%%%%%%%%
% !TEX root = MDPpaper.tex

\section{Results}
\label{sec:results}
%In this section, we present the details of the experiments carried out to demonstrate the effectiveness of our
%approach.
%We implement our approach as described in Section~\ref{sec:overview}.
%
We experimentally evaluate the research questions posed in the problem statement: \textit{question 1 -- given a
limited amount of data, can we use it efficiently to quantify a confidence that our system satisfies a given property?
question 2 -- can we design experiments that increase the accuracy of this confidence?}

\vspace{-0.3cm}
\subsubsection{Experimental Set-up.}
Our approach is implemented in C++.  % for computation in main memory, and run on
%an Intel(R) Xeon(R) E5-2440 12-Core 2.40GHz/64bit/96GB running Red Hat Linux
%6.3.1-1 with GCC 6.3.1. 
We use PRISM ~\cite{DBLP:conf/cav/KwiatkowskaNP11} for parameter synthesis, and 
GSL-2.3~\cite{DBLP:books/daglib/0026677} for random number generation. 

To answer question 2, we evaluate our \textit{synthesised strategy} approach 
against two alternatives. 
The first comparison is against a memoryless strategy, 
randomly selected from the set of all possible memoryless strategies. We term the resultant strategy as \textit{random
static strategy}. 
The second comparison strategy randomly selects actions at each state as data is collected, and therefore we term it as \textit{no strategy}. 
All three approaches use the same Bayesian inference framework over parameter counts. 

\begin{figure}[h]
\centering
\resizebox{0.6\textwidth}{!}{  
\begin{tikzpicture}[>=stealth',initial text={},shorten >=1pt,node distance=2.6cm,on grid,auto]
   \node[state,initial above] (0) {$S_{0}$};  
   \node[state,label={[align=center,xshift=1cm]below:\{complete\}}] (2) [below =of 0] {$S_{2}$};
   \node[state] (3) [right=of 2] {$S_{3}$};   
   \node[state] (4) [below left=of 0] {$S_{4}$};    
   \node[state] (1) [left =of 0] {$S_{1}$}; 
   \draw[name path=e0,->] (0) to node[xshift=-0.1cm,left] {$\frac{2}{5}$} (4);
   \draw[name path=e1,->] (0) to node {$(1-\param_1-\frac{1}{4})$} (3);
   \draw[name path=e2,<-] (0) to [loop right] node {$\frac{1}{4}$} (0);
   \draw[name path=e3,->] (0) to [bend left=30,pos=0.7] node {$\param_1$} (2);
   \draw[name path=e4,->] (2.80) to node [swap]{$1$} (0.280);   
   \draw[name path=e5,->] (0.260) to node [swap]{$1$} (2.100); 
   \draw[name path=e6,->] (0) to [bend right] node [swap]{$\frac{3}{5}$} (2);
   \draw[name path=e7,->] (0) to [bend right] node [swap]{$1$} (1);
   \draw[name path=e8,->] (1) to [bend right] node {$\frac{1}{10}$} (0);
   \draw[name path=e9,->] (1) to  [in=200,out=240,looseness=6] node [below,xshift=-0.2cm]{$\param_2$} (1);   
   \draw[name path=e10,->] (1.270) to [bend right=75,pos=0.6] node [swap,yshift=-0.3cm,xshift=1cm]{$(1-\param_2-\frac{1}{10})$} (2);
   \draw[name path=e11,->] (1.180) to [bend right=100, looseness=2] node [swap] {$1$} (2.250);
   \draw[name path=e12,->] (3) to [loop above] node {$1$} (3);   
   \draw[name path=e13,<-] (4) to [loop above] node {$1$} (4);
   \path[name path=c1] (0)  circle (0.8);
   \path[name path=c2] (0)  circle (0.8);
   \path[name path=c3] (1)  circle (0.8);
   \draw[name intersections={of=e0 and c1, by=i1},
     name intersections={of=e6 and c1, by=i2},
     red,-,shorten >=0pt]
     (i1) to [bend right] node {$ $} (i2);
    \draw[name intersections={of=e3 and c2, by=i3},
     name intersections={of=e2 and c2, by=i4},
     red,-,shorten >=6.5pt]
     (i3) to [bend right] node {$ $} (i4);
     \draw[name intersections={of=e8 and c3, by=i5},
     name intersections={of=e9 and c3, by=i6},
     red,-,shorten >=10pt]
     (i5) to [bend left] node {$ $} (i6); 
\end{tikzpicture}
}
\vspace{-25pt}
\caption{A simple pMDP 
%$\mathcal{M}^{p}$ 
for the experimental evaluation.} 
\label{fig:simple_PMDP}
\end{figure}
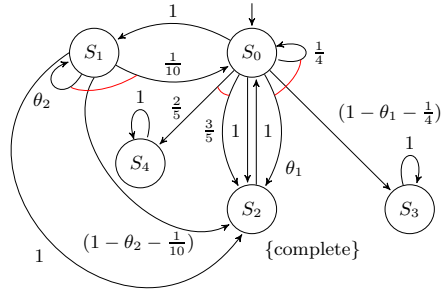

We present the analysis of our approach on the simple pMDP model in Fig.~\ref{fig:simple_PMDP} and with 
the PCTL property $\Props_{\geq 0.5}(\text{true}\,\,\mathcal{U}\,\,\textmd{complete})$. We also run our approach on models
up to 1000 states, but find the scalability depends on the number of actions in the model.
We assign non-informative priors to the parameters. %, i.e., Dirichlet distributions with hyperparameters $\mu = \{1,1\}$, for all $\param_i \in \param$.
%The allowable ranges for parameters $\param_1$ and $\param_2$ are $[0.0, 0.75]$ and  $[0.0, 0.9]$, respectively,  
%and their feasible sets are $[0.369, 0.75]$ and $[0.0, 0.9]$, respectively. 
Note that in our model, $\param_2$ does not contribute to the satisfaction of the property, 
and having validated that this does not affect the confidence results,
 we set $\param_2$ equal to $\param_1$. 
We simulate a range of underlying systems, 
corresponding to models $\M(\param)$ with different values for $\param$,  
which allows us 
to assess the accuracy of our confidence values against a ground truth, $G_{true}$.
For a simulated system modelled by $\M(\theta)$, this is given by: 
\begin{align}
\label{eq:trueres}
G_{true} =\left\{\begin{array}{ll}
    0       & \quad \text{if } \param_1 \notin [0.369,0.75],\\
    1  & \quad \text{if } \param_1 \in [0.369,0.75]. 
 \end{array}\right.
\end{align}

We collect data from the simulated system in the form of a history of state-action pairs visited.
We compute the mean squared error (MSE) between the ground truth from Eq.~\eqref{eq:trueres} and the confidence estimate, formally, \\
MSE $=
\frac{1}{n} \sum_{i=1}^{n} (G_{true} - G_i)^2$, where $n$ is the number of trials and $G_i$ is the output confidence
estimate for the $i$-th run.

\begin{figure}[h]
\centering
%\begin{subfigure}{.33\textwidth}
%  \centering
%  \resizebox{1.0\textwidth}{!}{
%  	\input{Plots/explicit_strategy_convergence_error_70.tex}
% %  } 
%  \caption{Synthesised strategy ($t$10,$l$10)}
%  \label{fig:118sub1}
%\end{subfigure}%
%\begin{subfigure}{.33\textwidth}
%  \centering
%  \resizebox{1.0\textwidth}{!}{
% 	 \input{Plots/random_memory_less_strategy_convergence_error_70.tex}
% %  }
%  \caption{Random static strategy ($t$10,$l$10)}
%  \label{fig:118sub2}
%\end{subfigure}
%\begin{subfigure}{.33\textwidth}
%  \centering
%  \resizebox{1.0\textwidth}{!}{
% 	 \input{Plots/random_finite_memory__convergence_error_70.tex}
% %  }
%  \caption{No strategy ($t$10,$l$10)}
%  \label{fig:118sub3}
%\end{subfigure}%
%\\ \vspace{5pt}
\begin{subfigure}{.33\textwidth}
  \centering
  \resizebox{1.0\textwidth}{!}{
\definecolor{mycolor1}{rgb}{1.00000,0.00000,1.00000}%
\begin{tikzpicture}

\begin{axis}[%
width=6.313in,
height=5.204in,
at={(1.059in,0.702in)},
scale only axis,
xmin=0.1,
xmax=0.8,
xtick={0.15,  0.2, 0.25,  0.3, 0.35,  0.4, 0.45,  0.5, 0.55,  0.6, 0.65,  0.7, 0.75},
xlabel style={font=\color{white!15!black}},
xlabel={System Parameter},
ymin=0,
ymax=0.45,
ylabel style={font=\color{white!15!black}},
ylabel={Mean Squared Error},
axis background/.style={fill=white},
legend style={legend cell align=left, align=left, draw=white!15!black}
]
\addplot [color=black, line width=1.0pt, mark=square, mark options={solid, black}]
  table[row sep=crcr]{%
0.15	0.0645339054979883\\
0.2	0.0615843767861007\\
0.25	0.136905412639165\\
0.3	0.193216916522028\\
0.35	0.269426798547559\\
0.4	0.297942897917525\\
0.45	0.224223738374969\\
0.5	0.195046791076965\\
0.55	0.121367184647797\\
0.6	0.0974413458962344\\
0.65	0.0666121202636401\\
0.7	0.05974443208782\\
0.75	0.0254589348030599\\
};
\addlegendentry{Synthesised Strategy}

\addplot [color=green, line width=1.0pt, mark=diamond*, mark options={solid, fill=green, green}]
  table[row sep=crcr]{%
0.15	0.12152185669455\\
0.2	0.171257042574814\\
0.25	0.231166131847321\\
0.3	0.211235036996757\\
0.35	0.280683518248554\\
0.4	0.307292966914627\\
0.45	0.272545983273459\\
0.5	0.280070732393258\\
0.55	0.250668003853206\\
0.6	0.15868763585576\\
0.65	0.178292022333572\\
0.7	0.11476377540388\\
0.75	0.09982656907941\\
};
\addlegendentry{Random Static Strategy}

\addplot [color=mycolor1, line width=1.0pt, mark=o, mark options={solid, mycolor1}]
  table[row sep=crcr]{%
0.15	0.1416759074\\
0.2	0.1927295997\\
0.25	0.2111938392\\
0.3	0.2389489032\\
0.35	0.2874263448\\
0.4	0.2967849903\\
0.45	0.3021256194\\
0.5	0.2370396794\\
0.55	0.1799355898\\
0.6	0.1824538623\\
0.65	0.1404674729\\
0.7	0.1113952385\\
0.75	0.0891556568999999\\
};
\addlegendentry{No Strategy}

\end{axis}
\end{tikzpicture}%
  } 
  \caption{All strategies ($t$10,$l$02)}
  \label{fig:118sub4}
\end{subfigure}%
\begin{subfigure}{.33\textwidth}
  \centering
  \resizebox{1.0\textwidth}{!}{
\definecolor{mycolor1}{rgb}{1.00000,0.00000,1.00000}%
\begin{tikzpicture}

\begin{axis}[%
width=6.313in,
height=5.204in,
at={(1.059in,0.702in)},
scale only axis,
xmin=0.1,
xmax=0.8,
xtick={0.15,  0.2, 0.25,  0.3, 0.35,  0.4, 0.45,  0.5, 0.55,  0.6, 0.65,  0.7, 0.75},
xlabel style={font=\color{white!15!black}},
xlabel={System Parameter},
ymin=0,
ymax=0.45,
ylabel style={font=\color{white!15!black}},
ylabel={Mean Squared Error},
axis background/.style={fill=white},
legend style={legend cell align=left, align=left, draw=white!15!black}
]
\addplot [color=black, line width=1.0pt, mark=square, mark options={solid, black}]
  table[row sep=crcr]{%
0.15	0.00475478650000005\\
0.2	0.0195304375000001\\
0.25	0.0872759315\\
0.3	0.1529579983\\
0.35	0.2133990444\\
0.4	0.4043204662\\
0.45	0.2370133368\\
0.5	0.1081211028\\
0.55	0.058250376\\
0.6	0.0252933926\\
0.65	0.0222674598\\
0.7	0.0167766749\\
0.75	0.0169215337\\
};
\addlegendentry{Synthesised Strategy}

\addplot [color=green, line width=1.0pt, mark=diamond*, mark options={solid, fill=green, green}]
  table[row sep=crcr]{%
0.15	0.0656600564000001\\
0.2	0.0821838913999999\\
0.25	0.1177076887\\
0.3	0.1897406441\\
0.35	0.2131144116\\
0.4	0.4222103265\\
0.45	0.2935986037\\
0.5	0.2741362403\\
0.55	0.1800180602\\
0.6	0.1633723618\\
0.65	0.0884576512\\
0.7	0.0625928597000001\\
0.75	0.017759479\\
};
\addlegendentry{Random Static Strategy}

\addplot [color=mycolor1, line width=1.0pt, mark=o, mark options={solid, mycolor1}]
  table[row sep=crcr]{%
0.15	0.059294795\\
0.2	0.1074491215\\
0.25	0.1526254435\\
0.3	0.215044892\\
0.35	0.3127001556\\
0.4	0.2705794253\\
0.45	0.2213226785\\
0.5	0.1393978117\\
0.55	0.1151918491\\
0.6	0.0695577856\\
0.65	0.0454914403\\
0.7	0.0338772541\\
0.75	0.016589945\\
};
\addlegendentry{No Strategy}

\end{axis}
\end{tikzpicture}%
  }
  \caption{All strategies ($t$10,$l$10)}
  \label{fig:118sub5}
\end{subfigure}
\begin{subfigure}{.33\textwidth}
  \centering
  \resizebox{1.0\textwidth}{!}{
\definecolor{mycolor1}{rgb}{0.00000,0.44700,0.74100}%
\definecolor{mycolor2}{rgb}{0.85000,0.32500,0.09800}%
\definecolor{mycolor3}{rgb}{0.92900,0.69400,0.12500}%
\definecolor{mycolor4}{rgb}{0.49400,0.18400,0.55600}%
\definecolor{mycolor5}{rgb}{0.46600,0.67400,0.18800}%
\definecolor{mycolor6}{rgb}{0.63500,0.07800,0.18400}%
\begin{tikzpicture}

\begin{axis}[%
width=6.313in,
height=5.204in,
at={(1.059in,0.702in)},
scale only axis,
xmin=0.1,
xmax=0.8,
xtick={0.15,  0.2, 0.25,  0.3, 0.35,  0.4, 0.45,  0.5, 0.55,  0.6, 0.65,  0.7, 0.75},
xlabel style={font=\color{white!15!black}},
xlabel={System Parameter},
ymin=0,
ymax=0.45,
ylabel style={font=\color{white!15!black}},
ylabel={Mean Squared Error},
axis background/.style={fill=white},
legend style={legend cell align=left, align=left, draw=white!15!black}
]
\addplot [color=mycolor1, line width=1.0pt, mark=o, mark options={solid, mycolor1}]
  table[row sep=crcr]{%
0.15	0.075390830938145\\
0.2	0.0899179121332496\\
0.25	0.143851310249334\\
0.3	0.160650242599683\\
0.35	0.251504870904027\\
0.4	0.378768002576585\\
0.45	0.256004344686236\\
0.5	0.22675507020983\\
0.55	0.169781516138835\\
0.6	0.133218923273313\\
0.65	0.0622224785774756\\
0.7	0.02939978526765\\
0.75	0.00110645527515996\\
};
\addlegendentry{02 Traces of Length 20}

\addplot [color=mycolor2, line width=1.0pt, mark=diamond, mark options={solid, mycolor2}]
  table[row sep=crcr]{%
0.15	0.048396653191218\\
0.2	0.0876029914880836\\
0.25	0.141966648320306\\
0.3	0.160512570015297\\
0.35	0.244006643327088\\
0.4	0.342410799395784\\
0.45	0.298819256128044\\
0.5	0.1631299442528\\
0.55	0.152508687465283\\
0.6	0.0413271128812299\\
0.65	0.02920657447332\\
0.7	0.00650710926610998\\
0.75	0.00393889650481005\\
};
\addlegendentry{04 Traces of Length 10}

\addplot [color=mycolor3, line width=1.0pt, mark=square, mark options={solid, mycolor3}]
  table[row sep=crcr]{%
0.15	0.0357664980312827\\
0.2	0.109793739205832\\
0.25	0.128150222843209\\
0.3	0.169610813716575\\
0.35	0.288070814880608\\
0.4	0.318471000062283\\
0.45	0.247801571230681\\
0.5	0.146532955366852\\
0.55	0.110163738169516\\
0.6	0.06021577311216\\
0.65	0.01807720404799\\
0.7	0.01320031609022\\
0.75	0.00254331277095998\\
};
\addlegendentry{05 Traces of Length 08}

\addplot [color=mycolor4, line width=1.0pt, mark=x, mark options={solid, mycolor4}]
  table[row sep=crcr]{%
0.15	0.032469712796412\\
0.2	0.0648642679205811\\
0.25	0.113514534836369\\
0.3	0.219446663060674\\
0.35	0.286928658817032\\
0.4	0.278754045800487\\
0.45	0.2269148889006\\
0.5	0.13530963358179\\
0.55	0.07173207351358\\
0.6	0.0514399277126923\\
0.65	0.0205600629518899\\
0.7	0.01425131863873\\
0.75	0.00436454010703002\\
};
\addlegendentry{08 Traces of Length 05}

\addplot [color=mycolor5, line width=1.0pt, mark=asterisk, mark options={solid, mycolor5}]
  table[row sep=crcr]{%
0.15	0.0645339054979883\\
0.2	0.0615843767861007\\
0.25	0.136905412639165\\
0.3	0.193216916522028\\
0.35	0.269426798547559\\
0.4	0.297942897917525\\
0.45	0.224223738374969\\
0.5	0.195046791076965\\
0.55	0.121367184647797\\
0.6	0.0974413458962344\\
0.65	0.0666121202636401\\
0.7	0.05974443208782\\
0.75	0.0254589348030599\\
};
\addlegendentry{10 Traces of Length 02}

\addplot [color=black, dashdotted, line width=1.0pt]
  table[row sep=crcr]{%
0.15	0.00475478650000005\\
0.2	0.0195304375000001\\
0.25	0.0872759315\\
0.3	0.1529579983\\
0.35	0.2133990444\\
0.4	0.4043204662\\
0.45	0.2370133368\\
0.5	0.1081211028\\
0.55	0.058250376\\
0.6	0.0252933926\\
0.65	0.0222674598\\
0.7	0.0167766749\\
0.75	0.0169215337\\
};
\addlegendentry{10 Traces of Length 10}

\addplot [color=mycolor6, line width=1.0pt, mark=triangle, mark options={solid, mycolor6}]
  table[row sep=crcr]{%
0.15	0.00689566380000006\\
0.2	0.00675032480000004\\
0.25	0.0209548163\\
0.3	0.0651017419000001\\
0.35	0.147910205\\
0.4	0.2656685903\\
0.45	0.1110476816\\
0.5	0.0544084692000001\\
0.55	0.0272048477\\
0.6	0.0168323419999999\\
0.65	0.0147644614\\
0.7	0.0144329784\\
0.75	0.0136625161\\
};
\addlegendentry{100 Traces of Length 10}

\end{axis}
\end{tikzpicture}%
  }
  \caption{Synthesised strategy}
  \label{fig:118sub6}
\end{subfigure}
\caption{Errors produced by the confidence computation for the three strategies considered. 
%Plots (a), (b) and (c) display the \textit{mean squared error} (MSE) produced for a set of 10 traces of length 10
% over an emulated underlying system with both parameters set to 0.7 while 
Plots (a) and (b) show the MSE for each type of strategy and for 10 traces of different trace lengths over different
simulated systems. 
Plot (c) presents the MSE for the \textit{synthesised strategy} over different simulated systems and combinations of number of traces with
varying trace lengths. }
\label{fig:118}
\end{figure}
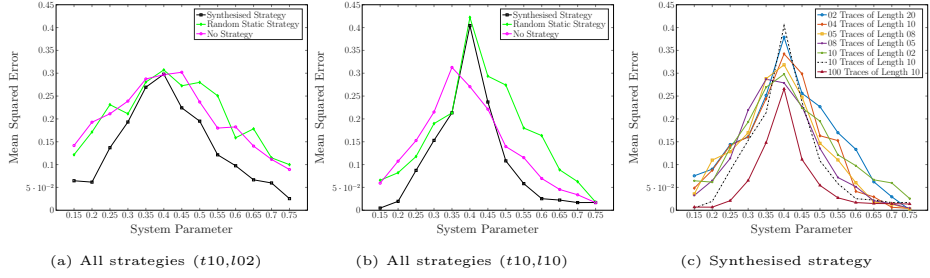

\begin{figure}
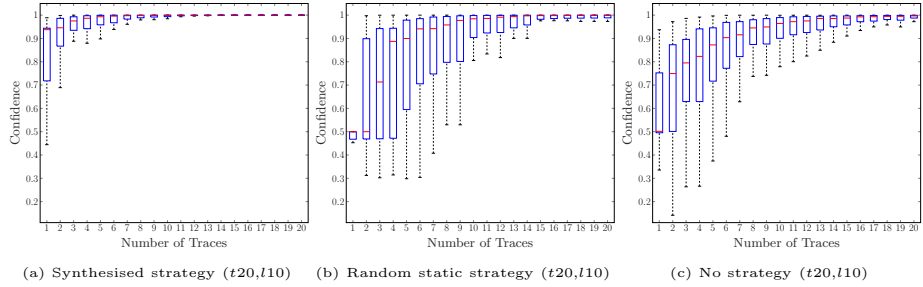

\centering
\begin{subfigure}{.33\textwidth}
  \centering
  \resizebox{1.0\textwidth}{!}{
  	\input{Plots/explicit_convergence_70p_20tr_10l.tex}
  }
  \caption{Synthesised strategy ($t$20,$l$10)}
  \label{fig:101sub1}
\end{subfigure}%
\begin{subfigure}{.33\textwidth}
  \centering
  \resizebox{1.0\textwidth}{!}{
  	\input{Plots/random_memoryless_convergence_70p_20tr_10l.tex}
  }
  \caption{Random static strategy ($t$20,$l$10)}
  \label{fig:101sub2}
\end{subfigure}%
\begin{subfigure}{.33\textwidth}
  \centering
  \resizebox{1.0\textwidth}{!}{
  	\input{Plots/random_finite_memory_convergence_70p_20tr_10l.tex}
  }
  \caption{No strategy ($t$20,$l$10)}
  \label{fig:101sub3}
\end{subfigure}%
%\\ \vspace{5pt}
%\begin{subfigure}{.33\textwidth}
%  \centering
%  \resizebox{1.0\textwidth}{!}{
% 	 \input{Plots/explicit_convergence_70p_10tr_50l.tex}
% %  }
%  \caption{Synthesised strategy ($t$10,$l$50)}
%  \label{fig:101sub4}
%\end{subfigure}
%\begin{subfigure}{.33\textwidth}
%  \centering
%%  \resizebox{1.0\textwidth}{!}{
% 	 \input{Plots/random_memoryless_convergence_70p_10tr_50l.tex}
% %  }
%  \caption{Random static strategy ($t$10,$l$50)}
%  \label{fig:101sub5}
%\end{subfigure}%
%\begin{subfigure}{.33\textwidth}
%  \centering
%  \resizebox{1.0\textwidth}{!}{
% 	 \input{Plots/random_finite_memory_convergence_70p_10tr_50l.tex}
% %  }
%  \caption{No strategy ($t$10,$l$50)}
%  \label{fig:101sub6}
%\end{subfigure}
\caption{Convergence of confidence outcomes to the ground truth over a simulated underlying system
with both parameters ($\param_1$, $\param_2$) set to be equal to 0.7. %, for traces of different trace lengths and each individual strategy considered.
}
\label{fig:101}
\vspace{-15pt}
\end{figure}

%\vspace{-30pt}
%
%
%
%\begin{center}
%\begin{figure}
%\centering
%\begin{subfigure}{.5\textwidth}
%  \centering
%  \includegraphics[width=0.95\linewidth]{parama.jpg}
%  \caption{Feasible sets for param `a'}
%  \label{fig:101sub1}
%\end{subfigure}%
%\begin{subfigure}{.5\textwidth}
%  \centering
%  \includegraphics[width=0.95\linewidth]{paramb.jpg}
%  \caption{Feasible sets for param `b'}
%  \label{fig:101sub2}
%\end{subfigure}
%\caption{Feasible sets for property  $\Props_{\geq 0.5}(true\,\,\mathcal{U}\,\,\textmd{``complete"})$ }
%\label{fig:101}
%\end{figure}
%\end{center}
%
%\begin{center}
%\begin{figure}
%\centering
% \input{Plots/param_synthesis.tex}
%\caption{Synthesised rectangular regions for property  $\Props_{\geq
%0.5}(true\,\,\mathcal{U}\,\,\textmd{``complete"})$. Areas coloured with \textit{red} indicate those regions of
%parameter evaluations that does not satisfy the property and areas coloured with \textit{green} indicate the feasible
%sets. Those areas in \textit{white} indicate regions that were undecided.} 
%\label{fig:101}
%\end{figure}
%\end{center}
%
\subsubsection{Observations and Discussion.}
The MSE in the confidence from all three strategies, 
over a range of underlying systems and varying quantities of data (i.e., for different numbers and lengths of traces), 
are shown in Fig.~\ref{fig:118}. The convergence of the confidence outcome is shown in Fig.~\ref{fig:101}, with
box plots showing the interquartile range (IQR), omitting any outliers, and whiskers extending to the most 
extreme data points not considered to be outliers. 

\paragraph{Accuracy of confidence results.}
The confidence for all approaches is low around the lower boundary of $\Param_{\prop}$, and the MSE is high, shown in Fig.~\ref{fig:118}. This is consistent with the
goal of the confidence calculation, 
where one would need to know the exact value of the system parameter $\theta$ if
its value is near this edge, to be able to decide whether it falls in $\Param_{\prop}$ or not, and hence the calculation has a high sensitivity around this boundary
 This sensitivity increases as the amount of data increases, as seen by comparing the MSE for $\param_1=0.4$ 
in Fig.~\ref{fig:118sub4}, where the
trace length is 2, with Fig.~\ref{fig:118sub5} when the trace length has increased up to 10.
To explore why this is the case, consider that to compute the confidence we integrate the posterior
distribution over the feasible set $\Param_{\prop} = [0.369, 0.75]$. 
The posterior distribution for $\param_i = 0.369$ should have a peak
centred at $0.369$ and half of the probability mass falling in the feasible set, leading to $\conf = 0.5$. 
The
height and width of the posterior distribution are determined by the amount and spread of data available and for 
a tall and thin distribution (encompassing a large amount of data), a small change in the position of the peak
can move a large percentage of mass of the distribution in or out of the feasible set.
This is prominent in Fig.~\ref{fig:118sub5} since our approach synthesises a
strategy that would yield the highest information gain, i.e., the most useful data. 
However,
as we move away from the edge, increased data effectively places probability mass away from the uncertain
regions, thus reducing both variance and MSE. 
Neither of the other two alternatives has the
ability to collect as much useful data and 
therefore variance is high even at the far ends of the parameter spectrum.
The ability of our method to collect more useful data 
is also illustrated in the 
convergence graphs shown in Fig.~\ref{fig:101}, where synthesis approach converges to the ground truth quicker than both comparison strategies.

We conclude that our strategy synthesis does improve the accuracy of the confidence calculation, 
unless the parameter value falls close to the boundary of $\Param_{\prop}$, and that away from this boundary the confidence converges to the ground truth and we are able to verify the property over $S$ based on the data collected.
%While answering question 1, we note that, with quantities of data approaching 100 traces of length
%10, all three strategies converge on a representation of the ground truth (with the exception of the case where the parameter value falls closer the boundary), and for the \textit{synthesised 
%strategy} this is evident even at lower amounts of data, see Fig.~\ref{fig:110sub1}, and we are able to thus verify the property over $S$ based on the data collected, demonstrating the data efficiency of the approach.

\paragraph{Robustness.} 
We run our implementation with varying lengths of traces, where the total number of transitions in the data remains the same, and the results summarised in Fig.~\ref{fig:118sub6} show that our approach, on this case study, is relatively insensitive
to this variation (compare Fig.~\ref{fig:118sub4} with Fig.~\ref{fig:118sub5}).
%outlines that our approach maintains a steady accuracy
%for much lesser amount of data and improves upon as more data is acquired. 
Our method depends on the number of parameterised transitions we visit and so depends on the trace
length being long enough to visit some parameterised transitions. This is in contrast 
to Statistical Model Checking techniques, where the accuracy of the approach depends on the trace length being
great enough to satisfy the property, e.g., to reach some desired state. In both cases this will vary
depending on the structure of the model.

%%%%%%%%%%%%%%%%%%%%%%%%%%%%%%%%%%%%%%%%%%%%
%Section 8: conclusions
%%%%%%%%%%%%%%%%%%%%%%%%%%%%%%%%%%%%%%%%%%%%
\vspace{-0.3cm}
\section{Conclusions}

In this paper, we have presented an approach for statistical verification of a fragment of 
unbounded-time PCTL properties
on partially unknown systems, by automating the design of smart experiments that 
maximise the amount of useful
data collected from the underlying system. 
We validate that our approach increases the accuracy of the
confidence that the system satisfies the property, compared
to selecting data randomly. We are able to achieve meaningful confidence outcomes with
comparably limited amounts of available data. 

%We plan to extend this work to non-linearly parameterised Markov decision
%processes, 
%and to use alternative parameter synthesis tools that allow handling 
%more parameters in the model and at faster run-times~\cite{Quatmann2016}.
We are pursuing
extensions of this framework for much wider class of probabilistic
models, in particular continuous time models, with a broad range of applications.

\bibliographystyle{splncs03}
\bibliography{pmdp_references} 
%\appendix
%\input{appendix.tex}

\end{document}